\documentclass{article} % For LaTeX2e
\usepackage{iclr2027_conference,times}

\newif\ifarxiv
\arxivtrue          % arXiv version
\ifarxiv
    \iclrfinalcopy
\fi

\usepackage{amsmath,amsfonts,bm}

\def\eqref#1{equation~\ref{#1}}
\def\1{\bm{1}}

\DeclareMathAlphabet{\mathsfit}{\encodingdefault}{\sfdefault}{m}{sl}
\SetMathAlphabet{\mathsfit}{bold}{\encodingdefault}{\sfdefault}{bx}{n}

\usepackage{hyperref}
\usepackage{url}

\title{Audio LLMs Know When They Can't Hear You}

\author{
Amirhosein Javadi
\thanks{Work done during an internship at Apple.}
\thanks{Corresponding authors: \texttt{amjavadi@ucsd.edu},
\texttt{m\_samraghrazlighi@apple.com}}
\textsuperscript{1,2},
Richa Dixit\textsuperscript{1},
Mehrdad Farajtabar\textsuperscript{1},
Minsik Cho\textsuperscript{1},
Devang Naik\textsuperscript{1},
\\
\bfseries
Mohammad Samragh\footnotemark[2]\textsuperscript{1}
\\[3pt]
\textsuperscript{1}Apple,
\textsuperscript{2}University of California San Diego
}

\usepackage{amsmath}
\usepackage{amssymb}
\usepackage{booktabs}
\usepackage{graphicx}
\usepackage{wrapfig}
\usepackage{caption}
\usepackage{booktabs}
\usepackage{graphicx}

\usepackage[textsize=tiny]{todonotes}
\begin{document}

\maketitle

\ifarxiv
    \lhead{Preprint. Under review}
\fi

\begin{abstract}
Audio large language models allow users to interact with the model through speech. 
When an input recording is too degraded, the model may misinterpret the user's query and respond based on an incorrect transcription.
In this paper, we study model-conditional transcription reliability: whether an Audio LLM can recognize when its own transcription is unreliable. 
We first prompt the Audio LLM to assess whether its own transcription would be reliable, and find that the model is a poor judge of its own transcription reliability: in most cases, it predicts that its transcription will be reliable.
We find that existing approaches, including speech quality predictors, audio LLM generation uncertainty, and transcript-conditioned WER estimation, provide limited signals for detecting transcription failures.
In contrast, we discover that transcription reliability is strongly represented in the model's audio-encoder representations. Based on this observation, we devise a lightweight reliability predictor that operates on representations extracted by the frozen audio encoder and predicts the reliability class before generation.
The reliability predictor can trigger a clarification request from the user when their voice query is predicted to be unreliable, while allowing reliable queries to proceed without modifying the underlying Audio LLM.
Our predictor achieves 81.10\% in-domain and 78.09\% cross-domain macro-F1 scores, outperforming the strongest baselines by 10.33 and 11.93 points, respectively. 
Finally, we show that reliability labels can transfer across Audio LLM families, and that transfer performance is closely related to the alignment of their model-specific reliability boundaries.
\end{abstract}
\section{Introduction}

Audio large language models (LLMs) allow users to interact with language models directly through speech, removing the need to first formulate a text query. This convenience, however, introduces a failure mode that does not arise in text-based interaction: before responding to the user's request, the model must first infer its linguistic content from the acoustic signal. When the recording is degraded by background noise, reverberation, distance from the microphone, or other acoustic conditions, this interpretation can fail. More importantly, the model may not recognize that it has failed. Instead, it can infer a plausible but incorrect query and confidently generate a response to that query. 

\begin{figure}[t]
    \centering
    \includegraphics[page=2, width=\linewidth]{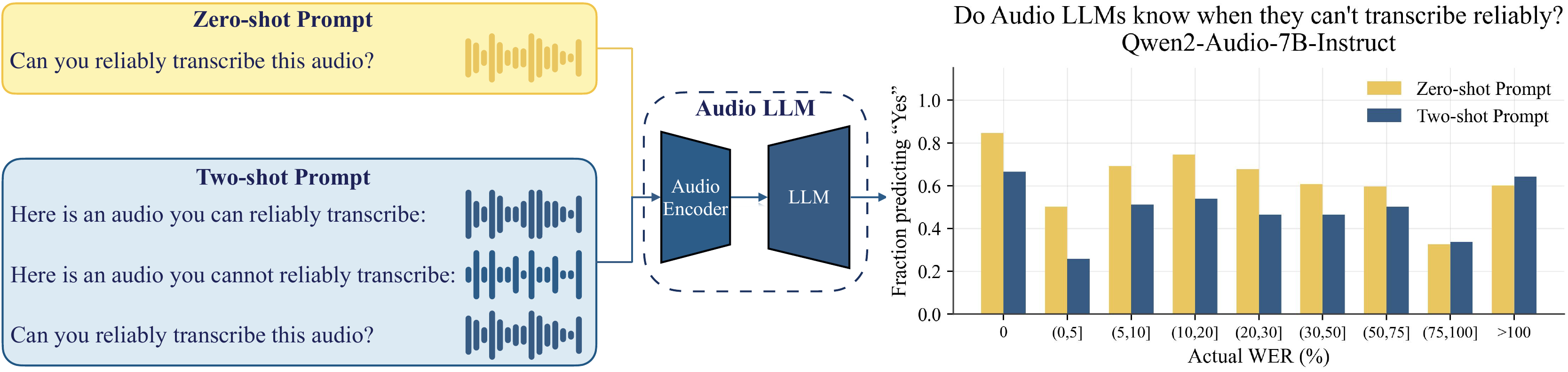}
    \vspace{-1mm}
    \caption{
    \textbf{Audio LLMs are poor judges of their own transcription reliability.}
    Bars show the probability of predicting reliable transcription within realized WER bins for acoustically corrupted speech. Both zero-shot and two-shot self-assessments are poorly aligned with actual reliability, with only modest improvement from two-shot prompting.
    }
    \vspace{-6mm}
    \label{fig:self_assessment}
\end{figure}

We raise a basic question for Audio LLMs: do they know when they cannot transcribe an input reliably? To answer this question and motivate the study, we begin with a simple self-assessment experiment in Figure~\ref{fig:self_assessment}. We progressively corrupt speech recordings, measure the actual transcription error of the Audio LLM, and separately ask the same model whether it expects its transcription of the recording to be reliable. We evaluate both zero-shot prompting and two-shot in-context learning on Qwen2-Audio-7B-Instruct \citep{chu2024qwen2audio}. 
Despite having access to the audio itself, the model's self-assessments frequently predict that its transcription will be reliable even for recordings on which it incurs substantial word error rate (WER).
Two-shot demonstrations reduce overconfidence in some cases, but do not completely resolve the problem. 

This observation suggests that the decoder in the language model lacks the ability to assess input quality before generation and cannot be reliably trusted with corrupted audio inputs.
% relying on the language-model decoder to recognize transcription failures may be insufficient. 
Modern Audio LLMs, however, contain another source of information: the pretrained audio encoder that transforms the input waveform into a sequence of acoustic representations before language generation. We find that these representations provide a considerably stronger signal of transcription reliability than the model’s generated-token probabilities. As such, rather than asking the decoder whether it understood the audio, we can directly predict transcription reliability from the audio encoder representations.

We therefore introduce a lightweight reliability predictor operating on top of a frozen Audio LLM encoder, which assigns the recording to one of four classes: reliable, minor degradation, moderate degradation, and severe degradation. When an input is predicted to be unreliable, the system can request clarification or ask the user to repeat the query rather than confidently responding to a potentially incorrect interpretation. Since the Audio LLM remains frozen, the reliability predictor can be incorporated without modifying the parameters of the audio encoder or language model.

To enable the training of our reliability predictor, we construct a transcription reliability dataset.
We additionally study whether model-grounded reliability labeling must be repeated for every Audio LLM.
We find that label banks can be reused across Audio LLM families, although the transfer gap varies across model pairs.
Moreover, we find that transfer degradation is closely related to how well the models' reliability boundaries align.
When a large transfer gap is associated with systematic differences in these boundaries, estimating and correcting boundary-specific shifts can substantially reduce the gap.

We evaluate our reliability predictor in two settings: (i) in-domain, where test samples are held out from the same training distribution, and (ii) cross-domain, where train and test samples come from different distributions. 
Across both evaluations, the proposed predictor substantially outperforms no-reference speech quality and intelligibility predictors \citep{tjandra2025audioboxaesthetics, reddy2021dnsmos, mittag2021nisqa, kumar2023squim}, Audio LLM generation uncertainty, and transcript-conditioned WER estimation \citep{park2025fast}. On Qwen2-Audio-7B-Instruct \citep{chu2024qwen2audio}, our predictor achieves 81.10\% in-domain and 78.09\% cross-domain macro-F1. The strongest baselines reach only 70.77\% and 66.16\%, respectively. 

In summary, our main contributions are:
\begin{itemize}
    \item We identify and systematically evaluate a self-assessment failure in Audio LLMs: models frequently remain confident that they can transcribe an input reliably even when their actual transcription error is high, and this behavior persists in the two-shot setting, where they are given one reliable and one unreliable audio example.

    \item We introduce model-grounded reliability labeling to construct a transcription reliability dataset. We show that frozen Audio LLM encoder representations support substantially more accurate reliability prediction than no-reference speech quality and intelligibility predictors, Audio LLM generation uncertainty, and transcript-conditioned WER estimation.
    
    \item We show that label banks can be reused across Audio LLM families, and that transfer degradation is closely related to how well their reliability boundaries align. When a large transfer gap is associated with systematic differences in these boundaries, correcting boundary-specific shifts can substantially reduce the gap.
    
\end{itemize}

\section{Related Work}
Prior work has studied several signals that may indicate whether speech can be transcribed reliably.
No-reference speech quality and intelligibility methods, including DNSMOS~\citep{reddy2021dnsmos}, NISQA~\citep{mittag2021nisqa}, TorchAudio-SQUIM~\citep{kumar2023squim}, and Audiobox-Aesthetics~\citep{tjandra2025audioboxaesthetics}, estimate perceptual quality, intelligibility, or related acoustic attributes directly from an audio recording.
These measures can provide useful proxies for transcription reliability, since degraded or unintelligible speech is more likely to lead to transcription errors.
However, the relationship is not exact: recordings with similar estimated quality can lead to different transcription outcomes depending on the utterance and the Audio LLM.
Our goal is therefore to predict whether a particular Audio LLM will transcribe a given recording reliably.

A more model-specific source of information comes from the Audio LLM's own transcription and generation process.
Audio LLM generation uncertainty can be estimated using signals such as token probabilities or predictive entropy over the generated transcription.
Related approaches estimate transcription error directly from the audio and the generated transcript.
For example, Fe-WER~\citep{park2025fast} combines acoustic representations of the input audio with textual representations of the generated transcript to estimate utterance-level WER without access to the reference transcript.
These methods provide model-dependent signals that are more closely tied to transcription errors, but they require language generation, either to obtain a transcript or to access uncertainty signals produced during decoding.
In contrast, we investigate whether transcription reliability can be predicted directly from the Audio LLM's audio-encoder representations without generating a transcript.

Beyond estimating reliability from generated outputs, prior work has also studied whether language models can recognize their own failures.
Prior approaches include verbalized uncertainty, where models express confidence in their answers~\citep{lin2022teaching}, and self-evaluation, where models assess the correctness of their own responses or whether they possess sufficient knowledge to answer~\citep{kadavath2022language}.
Similar challenges have been observed in multimodal models, where calibration and self-assessment remain difficult~\citep{chen2025unveiling}.
Our setting instead asks whether an Audio LLM can recognize when degraded speech will lead to an unreliable transcription.
We study this self-assessment capability and compare it with reliability prediction directly from the model's audio-encoder representations.

\begin{figure}[t]
    \centering
    \includegraphics[width=\linewidth]
    {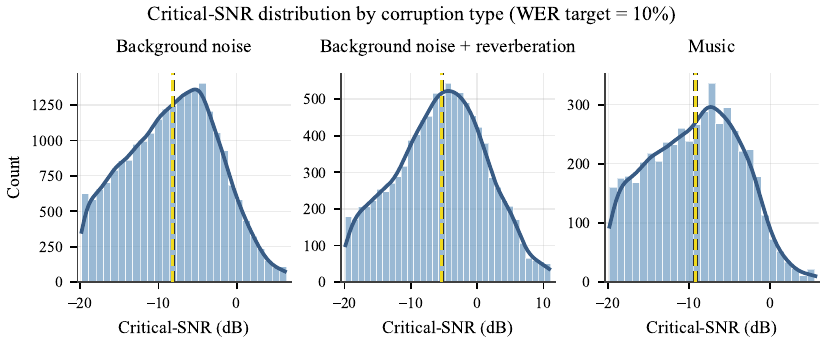}
    \caption{
    \textbf{Critical SNR varies substantially across speech--corruption pairs.}
    Distributions show the critical SNR required for Qwen2-Audio-7B-Instruct to achieve a WER no greater than $10\%$ under background noise, noise with reverberation, and music. The variation within each corruption category shows that a fixed SNR threshold cannot reliably characterize transcription reliability. Dashed lines indicate category means.
    }
    \vspace{-5mm}
    \label{fig:critical_snr}
\end{figure}
\section{Methodology}
\label{sec:Methodology}

Our goal is to predict whether a target Audio LLM can reliably transcribe a given audio recording. Our approach has two components. First, we construct a transcription reliability dataset through model-grounded reliability labeling by measuring how the transcription performance of the target Audio LLM changes under controlled acoustic corruption. Second, we train a lightweight reliability predictor on top of the frozen audio encoder using the resulting reliability labels.

\subsection{Audio LLM Reliability Dataset Construction}
\label{sec:data_construction}
Training the reliability predictor requires labels that reflect the transcription behavior of the target Audio LLM. We therefore construct a transcription reliability dataset through model-grounded reliability labeling, in which reliability boundaries are estimated directly from the model's transcription performance under controlled acoustic corruption.

For a speech utterance $x$, reference transcript $y$, and target Audio LLM $f(\cdot)$, we measure the word error rate (WER) as $E(f(x),y)$, where $E(\cdot,\cdot)$ denotes the transcription error. Given an additive interference signal $n$, the signal-to-noise ratio (SNR) is defined as
\begin{equation}
    \mathrm{SNR}(x,n)
    =
    20\log_{10}
    \left(
    \frac{\lVert x\rVert_2}{\lVert n\rVert_2}
    \right),
\end{equation}
where \(\lVert\cdot\rVert_2\) denotes the \(\ell_2\) norm.
For a desired SNR $s$, we scale the interference signal and define the resulting corrupted waveform as
\begin{equation}
    \widetilde{x}_s
    =
    x + \alpha_s n,
    \qquad
    \alpha_s
    =
    \frac{\lVert x\rVert_2}
         {\lVert n\rVert_2\,10^{s/20}}.
    \label{eq:snr_corruption}
\end{equation}
Since transcription error generally increases as SNR decreases, we define the critical SNR for model $f$ and WER threshold $\tau$ as

\begin{equation}
    s_{\tau,f}(x,n)
    =
    \min\left\{
    s :
    E\!\left(f(\widetilde{x}_s),y\right)
    \leq \tau
    \right\}.
    \label{eq:critical_snr}
\end{equation}
Thus, $s_{\tau,f}(x,n)$ is the lowest SNR at which model $f$ achieves a WER no greater than $\tau$ for the speech--corruption pair $(x,n)$.

Figure~\ref{fig:critical_snr} illustrates the distribution of critical SNR corresponding to a WER of $10\%$ across numerous speech--corruption pairs evaluated using Qwen2-Audio-7B-Instruct. Although the average critical SNR differs across corruption types, each category spans a wide range. Thus, even within the same corruption category, a single SNR threshold cannot reliably determine the transcription reliability of an input. This motivates estimating critical SNRs separately for each speech--corruption pair.

\begin{figure}[t]
    \centering
    \includegraphics[page=2, width=\linewidth]
    {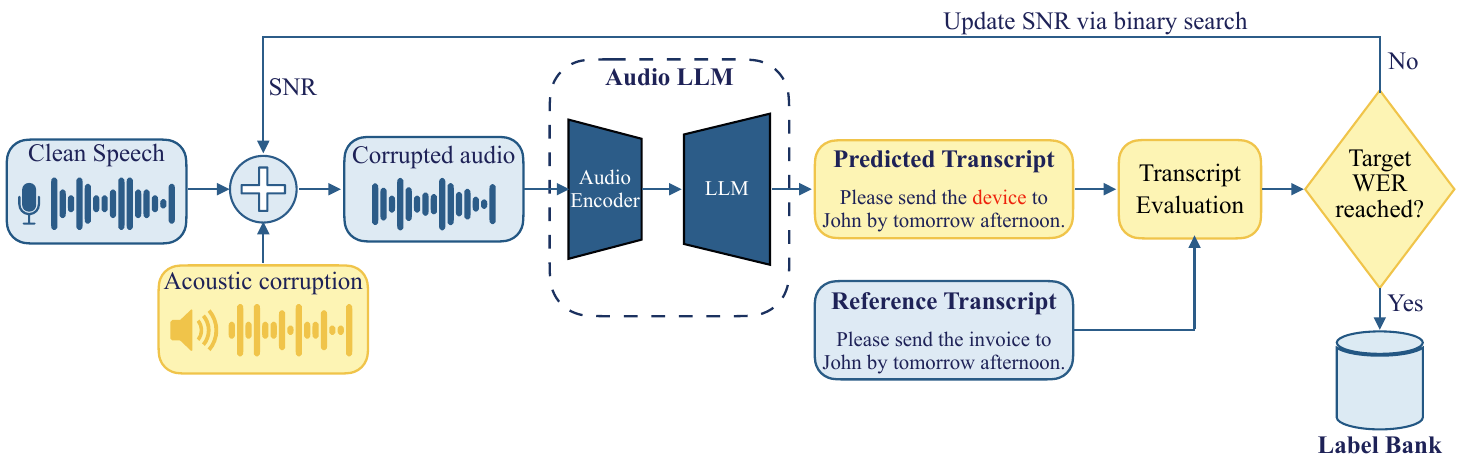}
    \caption{
    \textbf{Overview of the reliability dataset construction pipeline.}
    Starting from clean speech and a fixed acoustic corruption, we vary the
    corruption strength and obtain a transcription from the target Audio LLM.
    The generated transcript is compared with the reference transcript to
    compute WER. We then search for the corruption levels at which the model
    crosses the WER boundaries defining our reliability classes and store the
    resulting pair-specific boundaries in a label bank.
    }
    \vspace{-3mm}
    \label{fig:dataset_construction}
\end{figure}

In our experiments, we define four reliability classes: (i) reliable (WER \(=0\)), (ii) minor degradation (\(0 < \mathrm{WER} \leq 10\%\)), (iii) moderate degradation (\(10\% < \mathrm{WER} \leq 30\%\)), and (iv) severe degradation (\(\mathrm{WER} > 30\%\)). These classes are defined by the WER thresholds $\tau \in \{0.0,0.1,0.3\}$. For each speech--corruption pair, we estimate the corresponding critical SNRs and store them in a reliability label bank, referred to hereafter as the label bank.

Figure~\ref{fig:dataset_construction} summarizes the label-bank construction procedure. For each speech--corruption pair, we keep the speech utterance and corruption source fixed and vary only the SNR. Each resulting waveform is transcribed by the target Audio LLM and compared with the reference transcript to compute WER. Assuming an approximately monotonic relationship between SNR and transcription error, we use binary search to estimate the critical SNR associated with each WER threshold. Because these boundaries are estimated independently for every speech--corruption pair, the resulting label bank captures the transcription behavior of the target Audio LLM rather than relying on nominal corruption strength alone.

We consider both additive interference, including environmental noise and music, and reverberant conditions in which room reverberation is applied before additive interference. For each speech--corruption pair, the underlying speech, corruption source, and room impulse response, when applicable, are kept fixed while only the SNR is varied. 
To ensure high-quality reliability labels, we apply quality-control filtering before admitting a speech--corruption pair to the label bank. In particular, we exclude pairs for which the clean utterance is already transcribed poorly, reverberation alone causes substantial transcription error, the estimated critical SNRs are insufficiently separated or violate their expected ordering, or valid SNR intervals for the reliability classes cannot be identified. 

After filtering, training examples are generated by sampling SNRs within the retained class-specific SNR intervals. This allows the acoustic conditions to vary across training while preserving the reliability class assigned by the target Audio LLM. Validation and test examples are fixed so that all methods are evaluated on identical waveforms. Additional details on preprocessing, partitioning, critical-SNR estimation, quality-control criteria, and sample generation are provided in Appendix~\ref{app:data_construction}.
\subsection{Audio-Encoder Reliability Prediction}
\label{sec:model}

\begin{figure}[t]
    \centering
    \includegraphics[width=\linewidth]{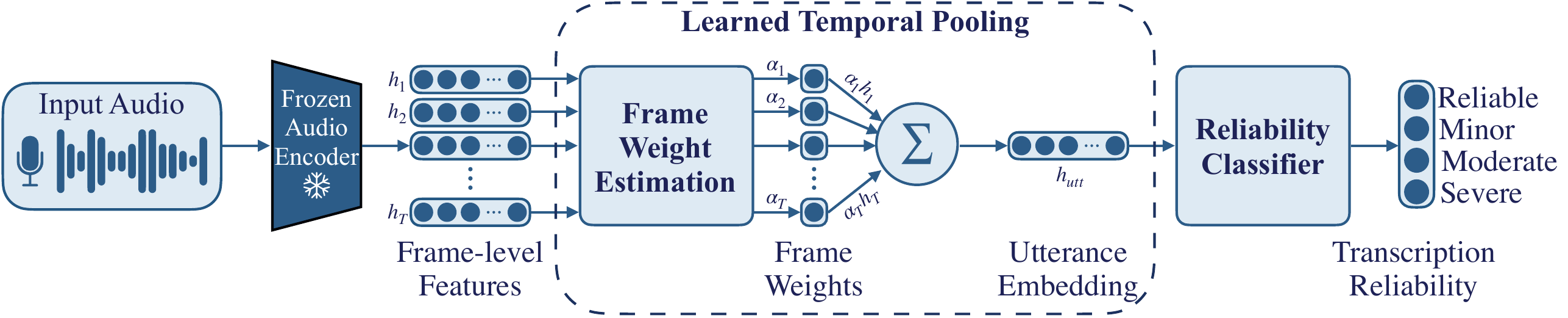}
    \caption{
    \textbf{Audio-encoder-based transcription reliability prediction.}
    The pretrained audio encoder is kept frozen and produces a sequence of frame-level representations. A learned temporal pooling module assigns a scalar weight to each frame and aggregates the weighted representations into a single utterance embedding. A lightweight classifier then predicts one of four reliability classes: reliable, minor degradation, moderate degradation, or severe degradation.
    }
    \vspace{-2mm}
    \label{fig:Model}
\end{figure}

Given an input audio clip and a target Audio LLM, our goal is to predict the reliability of the model's transcription without access to the reference transcript. Figure~\ref{fig:Model} illustrates the proposed architecture. We use the pretrained audio encoder of the target Audio LLM as the feature extractor and keep all of its parameters frozen. For an input waveform $x$, the encoder produces a sequence of frame-level representations, $\{h_1,\ldots,h_T\}$. Freezing the encoder preserves the representations learned during large-scale pretraining and ensures that training the reliability predictor does not modify the underlying Audio LLM.

The frame-level representations produced by the audio encoder are not necessarily equally informative for transcription reliability. We therefore use a lightweight learned temporal pooling module that assigns a scalar weight $\alpha_t$ to each frame representation $h_t \in \mathbb{R}^d$. For a sequence of $T$ vectors, $\alpha_t$ is computed using a learned linear projection followed by a Softmax operation:
\begin{equation}
    \alpha_t =
    \frac{\exp(w^\top h_t + b)}
    {\sum_{t'=1}^{T}\exp(w^\top h_{t'} + b)},
\end{equation}
where $w \in \mathbb{R}^d$ and $b \in \mathbb{R}$ are learned parameters. The aggregated sequence is summarized as
\begin{equation}
    h_{\mathrm{utt}} = \sum_{t=1}^{T} \alpha_t h_t .
\end{equation}
This produces a single utterance embedding $h_{\mathrm{utt}}$ while allowing the predictor to place different emphasis on different portions of the audio. The resulting embedding is then passed to a lightweight two-layer MLP that outputs logits for the four reliability classes.

The proposed predictor introduces negligible overhead compared with the standard Audio LLM inference pipeline. Since the audio encoder is already executed to process the input audio, the additional cost is limited to the lightweight pooling module and classification head. For example, in Qwen2-Audio-7B-Instruct, the predictor contains only 0.33M trainable parameters compared with the 636.97M frozen audio-encoder parameters (0.052\%), and adds approximately 2.12M FLOPs compared with the 1.89T FLOPs required by the audio encoder (0.00011\%). Detailed overhead comparisons across all evaluated Audio LLMs are provided in Appendix~\ref{app:efficiency}.
\section{Evaluation} 
\label{sec:evaluation}

\textbf{Datasets and evaluation setting.}
For the in-domain setting, we construct the transcription reliability dataset through model-grounded reliability labeling using LibriSpeech~\citep{panayotov2015librispeech} as the speech source, DNS Challenge~\citep{reddy2020dns} and MUSAN~\citep{snyder2015musan} as additive corruption sources, and simulated room impulse responses (RIRs) from OpenSLR26~\citep{ko2017dataaugmentation} for reverberation.
To evaluate generalization beyond the speech and acoustic conditions used during training, we additionally construct a cross-domain test set using LJSpeech~\citep{ljspeech17} for speech, SONYC-UST~\citep{cartwright2019sonyc} for environmental interference, and real room impulse responses from OpenSLR28~\citep{ko2017dataaugmentation}. The cross-domain test set is constructed using the same model-grounded reliability labeling procedure as the in-domain data, while all speech and acoustic sources are unseen during training.

\textbf{Implementation details.}
We freeze the pretrained audio encoder and train only the learned temporal pooling module and reliability classifier using cross-entropy loss. The classifier operates on the $d$-dimensional utterance embedding, where $d$ denotes the hidden dimension of the target Audio LLM's audio encoder. It consists of a linear layer from $d$ to 256 dimensions, followed by a GELU activation, dropout with probability 0.1, and a final linear layer producing logits for the four reliability classes. We optimize the trainable parameters with AdamW using a learning rate of $5\times10^{-5}$, a batch size of 256, and weight decay of $10^{-4}$. We train for 25 epochs with 3 warmup epochs and clip the gradient norm at 1.0. The best checkpoint is selected using validation macro-F1.

\subsection{Transcription Reliability Prediction}
\label{sec:main_results} 
Although our exact problem formulation has not been studied in prior work, we compare against the most closely related baselines as follows:

\textbf{Nominal SNR:} we treat the nominal mixing SNR as an oracle corruption-strength baseline; its exact value is available in our synthetically generated mixtures, but would generally not be accessible for real-world recordings where the clean speech and interference signals are not separately observed.

\textbf{No-reference speech quality and intelligibility predictors:} these methods assess audio quality by assigning a numeric value to a recording. We study Audiobox-Aesthetics~\citep{tjandra2025audioboxaesthetics}, DNSMOS~\citep{reddy2021dnsmos}, NISQA~\citep{mittag2021nisqa}, and TorchAudio-SQUIM~\citep{kumar2023squim}. For these methods, we use Production Quality (PQ), overall quality (OVRL), mean opinion score (MOS), and short-time objective intelligibility (STOI), respectively. 

\textbf{Audio LLM Generation Uncertainty:} we additionally evaluate two Audio LLM generation uncertainty measures from the target Audio LLM: \emph{mean token probability}, obtained by averaging the probabilities assigned to the generated tokens, and \emph{mean token entropy}, obtained by averaging predictive entropy across generation steps.

For the three scalar baseline families described above, we fit three ordered decision thresholds using only the in-domain training data to maximize four-class classification accuracy. These thresholds are then fixed for all evaluation sets.

\textbf{Transcript-conditioned WER estimation:} We study the method proposed in Fe-WER~\citep{park2025fast}, which estimates WER after decoding using an auxiliary model. The model is trained on our in-domain training data using the original architecture, with HuBERT-large~\citep{hsu2021hubert} for acoustic representations and XLM-R-large~\citep{conneau2020xlmr} for representations of the generated transcript. At evaluation time, the predicted WER is mapped directly to the same WER intervals used to define our four reliability classes. For the reliable class, which corresponds to zero WER, requiring a continuous regressor to predict exactly zero is overly restrictive. Since an utterance with $L$ reference words has a minimum nonzero WER of $\frac{1}{L}$, we assign predictions with $\widehat{\mathrm{WER}} < \frac{1}{L}$ to the reliable class and use the remaining fixed WER thresholds for the other classes. For trainable models, we use the in-domain validation set to select the best-performing checkpoint.

\begin{table}[t]
    \centering
    \caption{
    \textbf{Four-class transcription reliability prediction and cross-domain generalization for Qwen2-Audio-7B-Instruct.}
    For scalar baselines, three ordinal decision thresholds are calibrated
    using the training split and fixed during evaluation.
    Macro-F1, accuracy, and mean absolute error (MAE) are reported on the
    in-domain and cross-domain test sets.
    }
    \label{tab:main_results}

    \resizebox{\linewidth}{!}{
    \begin{tabular}{lcccccc}
    \toprule
    &
    \multicolumn{3}{c}{\textbf{In-domain Test}}
    &
    \multicolumn{3}{c}{\textbf{Cross-domain Test}}
    \\
    \cmidrule(lr){2-4}
    \cmidrule(lr){5-7}

    \textbf{Method}
    & \textbf{Macro-F1} $\uparrow$
    & \textbf{Accuracy} $\uparrow$
    & \textbf{MAE} $\downarrow$
    & \textbf{Macro-F1} $\uparrow$
    & \textbf{Accuracy} $\uparrow$
    & \textbf{MAE} $\downarrow$
    \\
    \midrule

    \multicolumn{7}{l}{\textit{Acoustic corruption proxy}} \\

    \quad SNR
    & 57.93
    & 65.14
    & 0.39
    & 51.59
    & 60.85
    & 0.47
    \\

    \midrule
    \multicolumn{7}{l}{\textit{No-reference speech quality and intelligibility predictors}} \\

    \quad Audiobox-Aesthetics (PQ)~\citep{tjandra2025audioboxaesthetics}
    & 22.43
    & 31.30
    & 1.10
    & 25.03
    & 34.26
    & 0.91
    \\

    \quad DNSMOS (OVRL)~\citep{reddy2021dnsmos}
    & 39.74
    & 47.88
    & 0.60
    & 41.76
    & 51.34
    & 0.59
    \\

    \quad NISQA (MOS)~\citep{mittag2021nisqa}
    & 45.74
    & 56.07
    & 0.48
    & 45.87
    & 61.88
    & 0.39
    \\

    \quad TorchAudio-SQUIM (STOI)~\citep{kumar2023squim}
    & 57.52
    & 62.48
    & 0.41
    & 55.04
    & 62.61
    & 0.40
    \\

    \midrule
    \multicolumn{7}{l}{\textit{Audio LLM generation uncertainty}} \\

    \quad Qwen mean token probability
    & 66.23
    & 69.24
    & 0.35
    & 66.16
    & 68.33
    & 0.36
    \\

    \quad Qwen mean token entropy
    & 66.73
    & 69.84
    & 0.34
    & 61.47
    & 63.25
    & 0.42
    \\

    \midrule
    \multicolumn{7}{l}{\textit{Transcript-conditioned WER estimation}} \\

    \quad Fe-WER~\citep{park2025fast}
    & 70.77
    & 75.05
    & 0.27
    & 64.67
    & 69.69
    & 0.33
    \\

    \midrule
    \multicolumn{7}{l}{\textit{Audio-encoder reliability prediction (ours)}} \\

    \quad \textbf{Qwen audio encoder + reliability predictor}
    & \textbf{81.10}
    & \textbf{82.60}
    & \textbf{0.18}
    & \textbf{78.09}
    & \textbf{79.55}
    & \textbf{0.22}
    \\

    \bottomrule
    \end{tabular}
    }
    \vspace{-3mm}
\end{table}

Table~\ref{tab:main_results} reports the main results using Qwen2-Audio-7B-Instruct~\citep{chu2024qwen2audio} as the target Audio LLM. Our reliability predictor achieves the strongest overall performance in both evaluation settings. On the in-domain test set, it achieves $82.60\%$ accuracy, $81.10\%$ macro-F1, and $0.18$ MAE. The strongest competing method, Fe-WER, reaches $75.05\%$ accuracy, $70.77\%$ macro-F1, and $0.27$ MAE. Our method therefore improves accuracy by $7.55$ percentage points and macro-F1 by $10.33$ points while reducing MAE by $0.09$.

Under cross-domain evaluation, our predictor retains $79.55\%$ accuracy and $78.09\%$ macro-F1 with an MAE of $0.22$. Among the baselines, Qwen mean token probability achieves the highest macro-F1 at $66.16\%$, while Fe-WER achieves the highest accuracy at $69.69\%$ and the lowest MAE at $0.33$. Relative to the strongest baseline for each metric, our method improves accuracy by $9.86$ points and macro-F1 by $11.93$ points while reducing MAE by $0.11$.

Importantly, the proposed predictor also exhibits a relatively small generalization gap: moving from the in-domain to the cross-domain test set reduces accuracy by only $3.05$ points and macro-F1 by $3.01$ points. In contrast, Fe-WER drops by $5.36$ accuracy points and $6.10$ macro-F1 points. This suggests that reliability information captured by the frozen Audio LLM encoder transfers more effectively across unseen speech, environmental interference, and room acoustics than transcript-conditioned WER estimation trained on the same in-domain data.

The acoustic corruption proxy and no-reference speech quality and intelligibility predictors are substantially weaker overall. Even nominal SNR, despite oracle access to the mixing SNR, reaches only $65.14\%$ in-domain accuracy and $60.85\%$ cross-domain accuracy. This supports the motivation in Section~\ref{sec:data_construction}: SNR alone does not fully determine the transcription behavior of an Audio LLM.

Finally, Audio LLM generation uncertainty provides a considerably stronger signal than the no-reference speech quality and intelligibility predictors, but remains substantially below our reliability predictor. This indicates that uncertainty in the generated token distribution is related to transcription failure but does not fully capture the reliability information already present in the model's audio-encoder representations. Our predictor also avoids language-model decoding: for Qwen2-Audio-7B-Instruct, the prediction head requires only $2.12$M FLOPs compared with $4.21$T FLOPs for the decoder, making the decoder roughly $2.0$ million times more computationally expensive than the added prediction head. Detailed comparisons are provided in Appendix~\ref{app:efficiency}.

\textbf{Additional studies:} We study two more Audio LLMs, namely Phi-4-Multimodal-Instruct~\citep{abouelenin2025phi4mini} and MOSS-Audio-8B~\citep{yang2026mossaudio}, in Appendix~\ref{app:additional_results}, where we follow the same baseline families and evaluation protocol. Appendix~\ref{app:temporal_aggregation} analyzes the temporal aggregation strategy used to summarize frame-level audio-encoder representations, while Appendix~\ref{app:encoder_probe_depth} studies the encoder depth at which those representations are probed. We also provide confusion-matrix analyses in Appendix~\ref{app:confusion_matrix} to examine error patterns across reliability classes.

\subsection{Cross-Model Transfer of Reliability Label Banks}
\label{sec:label_transfer}

\begin{figure}[t]
    \centering
    \includegraphics[width=\linewidth]{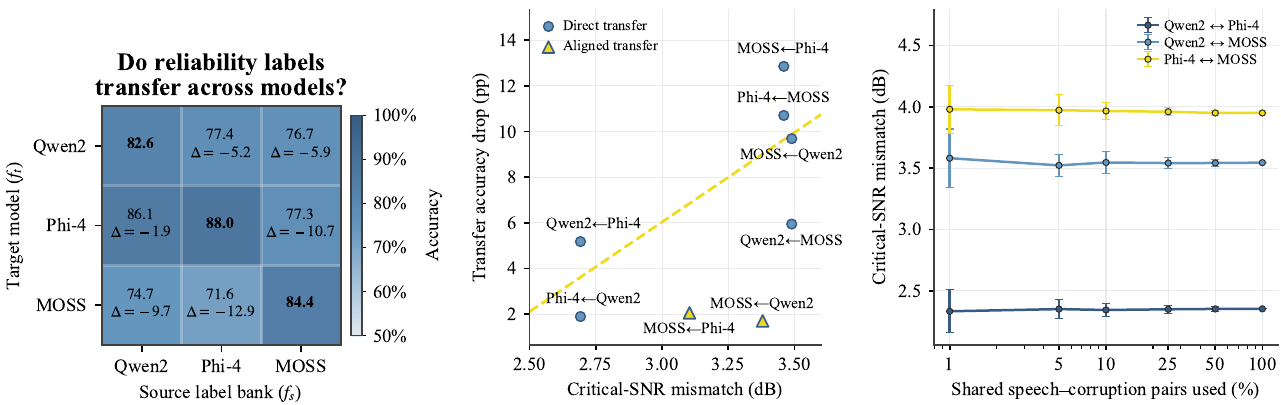}
    \caption{
    \textbf{Cross-model transfer of reliability label banks.}
    The left panel reports transfer accuracy when training a reliability predictor using label banks constructed by different Audio LLMs. The middle panel shows the relationship between critical-SNR mismatch and transfer degradation: blue circles show direct cross-model transfer, with larger mismatch associated with greater degradation, while yellow triangles show transfer after aligning the source critical-SNR boundaries using boundary-specific median shifts estimated from only $5\%$ of the shared speech--corruption pairs. This alignment substantially reduces the transfer gap for the two transfers to MOSS. The right panel evaluates the estimation of critical-SNR mismatch from different fractions of the shared pairs, reporting the mean and standard deviation over 20 random subsets. The estimate approaches the full-label-bank value even when using only a small fraction of the shared pairs.
    }
    \vspace{-5mm}
    \label{fig:label_transfer}
\end{figure}

Constructing a reliability label bank requires repeatedly evaluating the target Audio LLM to estimate critical SNRs across speech--corruption pairs and WER thresholds. We therefore investigate whether a label bank constructed for one Audio LLM can be reused to train a reliability predictor for another, thereby reducing the cost of constructing a complete model-specific label bank. We study this effect among three Audio LLMs: Qwen2-Audio-7B-Instruct (Qwen2), Phi-4-Multimodal-Instruct (Phi-4), and MOSS-Audio-8B (MOSS).

The left panel of Figure~\ref{fig:label_transfer} shows the performance of the reliability classifier for the Audio LLMs under study. The diagonal entries show the performance when the target model itself is used to construct the training label bank. We refer to the off-diagonal entries as \emph{cross-model transfer accuracy}, as they show the performance when the label bank constructed using one model is used to train a reliability predictor for another.

We make two observations. First, transfer between Qwen2 and Phi-4 is effective, largely preserving accuracy. In contrast, transfer between MOSS and either of the other two models results in a substantially larger degradation in accuracy. This raises the question: why does transfer work better between Qwen2 and Phi-4 than between MOSS and the other models? We hypothesize that this variation in transfer performance is related to the mismatch between the critical-SNR boundaries of the source and target models. Using the critical-SNR definition in Eq.~\ref{eq:critical_snr}, we define the \emph{critical-SNR mismatch} between a source model $f_s$ and a target model $f_t$ as
\begin{equation}
\label{eq:critical_snr_mismatch}
D(f_s,f_t)
=
\frac{1}{|\mathcal{P}||\mathcal{T}|}
\sum_{(x,n)\in\mathcal{P}}
\sum_{\tau\in\mathcal{T}}
\left|
s_{\tau,f_s}(x,n)
-
s_{\tau,f_t}(x,n)
\right|,
\end{equation}
where $\mathcal{P}$ denotes the shared speech--corruption pairs and $\mathcal{T}$ denotes the WER thresholds defining the reliability classes. The middle panel of Figure~\ref{fig:label_transfer} compares the critical-SNR mismatch with the corresponding accuracy drop under cross-model transfer. 
Each blue circle represents a transfer from source model $f_s$ to target model $f_t$, denoted by $f_t \leftarrow f_s$.
We observe that larger differences between the models' critical-SNR boundaries are associated with greater transfer degradation.

The right panel of Figure~\ref{fig:label_transfer} examines whether the critical-SNR mismatch can be estimated without constructing the complete target label bank. For each fraction of shared speech--corruption pairs, we compute the mismatch using the corresponding critical SNRs from the source and target models. We repeat this random subsampling procedure 20 times and report the mean and standard deviation. Even with a small fraction of the shared pairs, the estimated mismatch is close to the value from the full label bank, while its variance decreases as more pairs are used. This suggests that the critical-SNR mismatch can be estimated from a relatively small number of measurements.

Motivated by this observation, we estimate the shift between the source and target critical-SNR boundaries using only a sampled fraction of the shared speech--corruption pairs. Let $\mathcal{P}_{\rho}\subset\mathcal{P}$ denote a subset containing a fraction $\rho$ of the shared pairs. For transfer from $f_s$ to $f_t$, we estimate a separate shift for each WER threshold $\tau\in\mathcal{T}$ as
\begin{equation}
    \label{eq:critical_snr_shift}
    \delta_{s\rightarrow t,\tau}^{(\rho)}
    =
    \operatorname{median}_{(x,n)\in\mathcal{P}_{\rho}}
    \left[
    s_{\tau,f_t}(x,n)
    -
    s_{\tau,f_s}(x,n)
    \right].
\end{equation}
Here, $\delta_{s\rightarrow t,\tau}^{(\rho)}$ estimates the systematic shift between the source and target models for reliability boundary $\tau$. We use these estimated shifts to adjust the critical-SNR values of the source label bank to approximate those of the target model. Specifically, we align each source critical-SNR as
\begin{equation}
    \label{eq:shifted_critical_snr}
    \widehat{s}_{\tau,f_t}(x,n)
    =
    s_{\tau,f_s}(x,n)
    +
    \delta_{s\rightarrow t,\tau}^{(\rho)},
    \qquad
    \forall \tau \in \mathcal{T}.
\end{equation}

Using these shifted critical SNRs improves transfer accuracy. We show two examples as the yellow triangles in the middle panel of Figure~\ref{fig:label_transfer}, where the Phi-4 and Qwen2 label banks are adjusted for transfer to MOSS. The alignment uses only $5\%$ of the shared speech--corruption pairs to estimate the boundary-specific shifts between the source and target critical-SNR boundaries. For MOSS$\leftarrow$Phi-4, the accuracy drop decreases from $12.9$ to $2.1$ points, while for MOSS$\leftarrow$Qwen2, it decreases from $9.7$ to $1.7$ points. By using only $5\%$ of the shared pairs, we can estimate the boundary-specific shifts that largely close the performance gap introduced by direct cross-model transfer, substantially reducing the need for target-specific label construction.

\textbf{Additional study:}
We provide an ablation of the critical-SNR alignment strategy in Appendix~\ref{app:shift_strategy_ablation}, comparing global and per-boundary shift estimation using mean and median aggregation.
\section{Conclusion}

We studied whether Audio LLMs can recognize when their own transcription of an input recording is unreliable. Our self-assessment experiments show that current Audio LLMs remain poorly calibrated to their actual transcription error. At the same time, we find that transcription reliability is strongly encoded in the audio-encoder representations. Based on this observation, we introduced a lightweight reliability predictor trained on top of a frozen Audio LLM encoder using a dataset constructed through model-grounded reliability labeling. Our labeling procedure estimates pair-specific critical SNRs from the measured transcription behavior of the target Audio LLM. The resulting predictor substantially outperforms no-reference speech quality and intelligibility predictors, Audio LLM generation uncertainty, and transcript-conditioned WER estimation, while maintaining strong performance under cross-domain evaluation.

We further showed that reliability label banks can be reused across Audio LLM families, although transfer performance depends on the source--target model pair. Transfer degradation is closely associated with the critical-SNR mismatch between models, and boundary-specific shifts in their critical-SNR boundaries can be estimated to reduce large transfer gaps. Together, these results suggest that Audio LLMs contain useful information about when their transcriptions are likely to fail, even when that information is not reliably expressed through their generated outputs. Explicitly exposing this signal provides a practical mechanism for detecting unreliable audio queries before generation and enabling systems to request clarification rather than confidently responding to a misinterpreted input.

\ifarxiv

\section*{Acknowledgments}
We thank Dr. Masood Delfarah for insightful discussions and valuable feedback on this work.

\else

\section*{Reproducibility Statement}
We provide detailed information to support reproducibility of our experiments throughout the main paper and Appendix. Section~\ref{app:data_construction} describes the data sources, preprocessing, train--validation--test partitioning, acoustic corruption procedure, critical-SNR estimation, quality-control criteria, and training and evaluation sample generation. The main paper specifies the reliability predictor architecture, optimization settings, and evaluation protocol.

\section*{AI Use Statement}
Generative AI tools were used to assist with grammar editing and to improve the clarity and organization of the manuscript. They were not used to generate the original text, experimental results, fabricate data, or replace the authors' scientific judgment. All technical content, experimental design, analyses, reported results, and conclusions were developed, verified, and approved by the authors, who take full responsibility for the contents of the paper.

\fi

\bibliography{iclr2027_conference}
\bibliographystyle{iclr2027_conference}

\appendix
\clearpage

\section{Appendix Overview}

In this section, we provide a brief overview of the Appendix contents. The Appendix provides additional analyses, implementation details, and ablation studies that complement the main paper.

Section~\ref{app:self_assessment} provides additional details on the Audio LLM self-assessment experiments, including the zero-shot and two-shot prompting settings. 
Section~\ref{app:data_construction} describes the reliability label construction pipeline, including data preprocessing, critical SNR estimation, quality control, and sample generation procedures. 
Section~\ref{app:additional_results} provides additional transcription reliability prediction results for Phi-4-Multimodal-Instruct and MOSS-Audio-8B. 
Section~\ref{app:confusion_matrix} presents additional confusion matrix analyses for reliability prediction across different Audio LLMs. 
Section~\ref{app:efficiency} analyzes the computational overhead of the proposed reliability predictor and quantifies its parameter and FLOPs cost. 
Finally, Section~\ref{app:ablation} investigates temporal aggregation strategies, audio-encoder probe depth, and the critical-SNR alignment strategy.

\section{Audio LLM Self-Assessment Details}
\label{app:self_assessment}

We evaluate whether an Audio LLM can predict the reliability of its own transcription of an input recording. 
For each target Audio LLM, we generate noisy versions of held-out speech recordings by varying the signal-to-noise ratio (SNR) from $-20$ to $20$~dB. For each resulting recording, we first ask the model to transcribe the audio and compute its WER against the reference transcript. We then query the model separately about whether it expects its own transcription of the same recording to be reliable. The reference transcript is used only to compute the actual WER for evaluation and is never provided to the model during self-assessment. We evaluate this capability under both zero-shot prompting and two-shot in-context learning (ICL).

\textbf{Zero-shot self-assessment.} In the zero-shot setting, the model receives only the test recording together with the definition of transcription reliability. It is asked to predict whether its own transcription of that recording would be reliable, without explicitly transcribing the audio. The model is constrained to answer with either \texttt{Yes} or \texttt{No}. We use the following prompt:

\begin{quote}
\small
You will hear a speech recording. Your task is to predict whether your
own transcription of this exact recording would be reliable.

For this task, ``reliable'' means that if you transcribed the recording
now, your word error rate (WER) would be exactly $0\%$ relative to the
correct transcript. WER counts word substitutions, deletions, and
insertions.

\medskip
\textbf{Test recording:} \texttt{\{test\}}

\medskip
Based only on the audio, predict whether your own transcription would
satisfy this criterion. Do not transcribe the recording.

Answer only \texttt{Yes} or \texttt{No}.
\end{quote}

\textbf{Two-shot self-assessment.}
For the two-shot setting, we additionally provide one reliable and one unreliable audio example before the test recording. To construct these examples, we first build a model-specific context pool containing 100 speech utterances. Each utterance is required to have zero WER when transcribed by the target model in the clean condition. For each utterance, we then use the same binary-search procedure as in our dataset construction to estimate two critical SNRs: the lowest SNR at which the model achieves zero WER and the lowest SNR at which the model achieves a WER of $30\%$ or lower. The corrupted recording at the critical SNR for zero WER is used as the reliable demonstration, while the corrupted recording at the critical SNR for $30\%$ WER is used as the unreliable demonstration.

For each test recording and noise instance, we sample five different utterances from this context pool and query the model five separate times. Each query contains exactly one reliable demonstration, one unreliable demonstration, and the test recording. The final self-assessment for the test recording is determined by majority vote over the five responses.

Each of the five model queries uses the following prompt:

\begin{quote}
\small
You will hear a speech recording. Your task is to predict whether your
own transcription of this exact recording would be reliable.

For this task, ``reliable'' means that if you transcribed the recording
now, your word error rate (WER) would be exactly $0\%$ relative to the
correct transcript. WER counts word substitutions, deletions, and
insertions.

\medskip
\textbf{Example 1:} \texttt{\{yes\}}\\
\textbf{Answer:} \texttt{Yes}

\medskip
\textbf{Example 2:} \texttt{\{no\}}\\
\textbf{Answer:} \texttt{No}

\medskip
\textbf{Test recording:} \texttt{\{test\}}

\medskip
Based only on the audio, predict whether your own transcription would
satisfy this criterion. Do not transcribe the recording.

Answer only \texttt{Yes} or \texttt{No}.
\end{quote}

\section{Dataset Construction Details}
\label{app:data_construction}

This section provides additional details for the model-grounded reliability labeled dataset construction described in Section~\ref{sec:data_construction}. The key principle is that reliability labels are determined by the measured transcription performance of the target Audio LLM, rather than by the acoustic corruption level itself. The same SNR can lead to substantially different transcription errors across speech utterances, corruption sources, and target models. We therefore estimate critical SNRs independently for each speech--corruption pair.

\paragraph{Data sources and preprocessing.}
For the in-domain experiments, we use LibriSpeech as the source of clean speech, with train-clean-100, dev-clean, and test-clean used for training, validation, and testing, respectively. All recordings are converted to mono audio at $16$~kHz, and we retain utterances between $2$ and $30$ seconds. Before constructing corrupted examples, each clean utterance is transcribed by the target Audio LLM, and examples with insufficient clean-speech transcription accuracy (WER $\geq 5\%$) are removed. This prevents errors already present in the clean recording from being attributed to acoustic degradation.

Speech and acoustic corruption sources are partitioned before label construction. The standard LibriSpeech splits provide disjoint speakers and utterances, while DNS and MUSAN recordings are partitioned into $70\%/10\%/20\%$ training, validation, and test subsets. OpenSLR26 RIRs are partitioned at the room level so that different measurements from the same acoustic environment cannot appear across data splits.

For cross-domain evaluation, we replace all three acoustic components with sources that are unseen during training. We use LJSpeech for clean speech, SONYC-UST for environmental interference, and real room impulse responses from OpenSLR28. The cross-domain test set is constructed using the same model-grounded reliability labeling procedure as the in-domain test set.

\paragraph{Acoustic corruption.}
For each clean speech utterance, we construct examples using either additive interference alone or additive interference after applying room reverberation. The interference signal is scaled to obtain the desired SNR and then mixed with the speech waveform. For a given speech--corruption pair, the speech recording, corruption source, and RIR, when applicable, are kept fixed while the SNR is varied. This allows us to measure how the transcription of the target Audio LLM changes as only the strength of the acoustic degradation is varied.

\paragraph{Critical SNR estimation.}
For each speech--corruption pair, we vary the SNR and transcribe the resulting audio with the target Audio LLM. We search over SNRs from $-20$ to $30$~dB and use binary search to estimate the critical SNR associated with each target WER. For a target WER $\tau$, the corresponding critical SNR is the lowest SNR at which the target Audio LLM achieves $\mathrm{WER} \leq \tau$. Binary search is performed with a tolerance of $0.5$~dB and a maximum of 15 iterations. If a required critical SNR is not reached within the search range, it is treated as undefined and the corresponding reliability class is not sampled for that speech--corruption pair. Importantly, the critical SNRs are estimated independently for every speech--corruption pair and every target Audio LLM. Transcription is performed using deterministic decoding. 

In our experiments, the target WERs define four reliability classes: \emph{reliable}, corresponding to WER $=0$; \emph{minor degradation}, corresponding to $0 < \mathrm{WER} \leq 10\%$; \emph{moderate degradation}, corresponding to $10\% < \mathrm{WER} \leq 30\%$; and \emph{severe degradation}, corresponding to $\mathrm{WER} > 30\%$. Consequently, examples assigned to the same reliability class need not have similar absolute SNRs: a particular SNR may be fully transcribable for one speech--corruption pair and unreliable for another.

\paragraph{Quality control and label-bank construction.}
To ensure high-quality reliability labels, we apply quality-control filtering before admitting a speech--corruption pair to the label bank. We remove cases where the clean recording is already transcribed poorly (WER $\geq 5\%$), reverberation alone causes substantial error (WER $\geq 5\%$), the estimated critical SNRs are too close to support stable sampling (less than $1$~dB apart), or the estimated critical-SNR boundaries violate their expected ordering. Table~\ref{tab:filtering_stats} summarizes the effect of these filtering criteria across the three target Audio LLMs. The majority of candidate speech--corruption pairs are retained, while low-quality or ambiguous cases are filtered out before label-bank construction.

To further reduce ambiguity, we exclude a small margin around the estimated critical SNRs shared by neighboring reliability classes when sampling examples. Specifically, for each class-specific SNR interval, we remove $5\%$ of that interval's width adjacent to each shared critical-SNR boundary. The retained information is stored in the label bank rather than as pre-generated noisy waveforms. Each entry contains the clean speech utterance, corruption source, the RIR when applicable, and the model-specific critical SNRs that define the class-specific SNR intervals.

\begin{table}[t]
    \centering
    \caption{
    \textbf{Effect of quality-control filtering on candidate speech--corruption pairs.}
    We report the percentage of candidate pairs removed by each filtering criterion for each target Audio LLM, together with the fraction retained for label-bank construction.
    }
    \label{tab:filtering_stats}
    \vspace{1mm}

    \setlength{\tabcolsep}{2.3pt}
    \normalsize
    \begin{tabular}{@{}lccccc@{}}
        \toprule
        Model
        & Clean Fail.
        & Reverb Fail.
        & SNR Gap
        & Order Viol.
        & Retained \\
        \midrule

        Qwen2-Audio-7B-Instruct
        & 3.8\%
        & 1.6\%
        & 7.3\%
        & 4.0\%
        & 83.3\% \\

        Phi-4-Multimodal-Instruct
        & 2.1\%
        & 1.0\%
        & 3.8\%
        & 5.1\%
        & 88.1\% \\

        MOSS-Audio-8B
        & 8.5\%
        & 4.0\%
        & 3.4\%
        & 4.4\%
        & 79.7\% \\

        \bottomrule
    \end{tabular}
\end{table}

\paragraph{Training and evaluation sample generation.}
During training, a reliability class is first selected and an SNR is sampled uniformly at random from the corresponding class-specific SNR interval of the stored speech--corruption pair. The SNR is resampled across epochs, allowing the same underlying speech--corruption pair to produce different acoustic realizations while preserving its reliability label. For the reliable class, $40\%$ of examples use the clean speech recording, while the remaining $60\%$ use corrupted speech for which the target Audio LLM retains zero WER. Validation and test examples are instead instantiated once and kept fixed across all experiments. For each retained speech--corruption pair and available reliability class, we select a fixed SNR from within the corresponding class-specific SNR interval and use the same resulting waveform for every method being evaluated.

\begin{table}[t]
    \centering
    \caption{
    \textbf{Four-class transcription reliability prediction and cross-domain generalization for Phi-4-Multimodal-Instruct.}
    We report macro-F1, accuracy, and mean absolute error (MAE) on the in-domain and cross-domain test sets, following the same baselines and evaluation protocol as the Qwen2-Audio-7B-Instruct results in Table~\ref{tab:main_results}. The proposed reliability predictor achieves the strongest overall performance in both settings.
    }
    \label{tab:phi4_main_results}

    \resizebox{\linewidth}{!}{
    \begin{tabular}{lcccccc}
    \toprule
    &
    \multicolumn{3}{c}{\textbf{In-domain Test}}
    &
    \multicolumn{3}{c}{\textbf{Cross-domain Test}}
    \\
    \cmidrule(lr){2-4}
    \cmidrule(lr){5-7}

    \textbf{Method}
    & \textbf{Macro-F1} $\uparrow$
    & \textbf{Accuracy} $\uparrow$
    & \textbf{MAE} $\downarrow$
    & \textbf{Macro-F1} $\uparrow$
    & \textbf{Accuracy} $\uparrow$
    & \textbf{MAE} $\downarrow$
    \\
    \midrule

    \multicolumn{7}{l}{\textit{Acoustic corruption proxy}} \\

    \quad SNR
    & 60.72 & 68.38 & 0.36
    & 54.39 & 64.09 & 0.43
    \\

    \midrule
    \multicolumn{7}{l}{\textit{No-reference speech quality and intelligibility predictors}} \\

    \quad Audiobox-Aesthetics (PQ)~\citep{tjandra2025audioboxaesthetics}
    & 23.01 & 32.88 & 1.12
    & 27.38 & 36.70 & 0.94
    \\

    \quad DNSMOS (OVRL)~\citep{reddy2021dnsmos}
    & 38.14 & 45.43 & 0.64
    & 41.05 & 49.57 & 0.64
    \\

    \quad NISQA (MOS)~\citep{mittag2021nisqa}
    & 52.46 & 57.36 & 0.47
    & 48.97 & 61.41 & 0.39 
    \\

    \quad TorchAudio-SQUIM (STOI)~\citep{kumar2023squim}
    & 57.92 & 62.37 & 0.40
    & 54.96 & 62.03 & 0.40
    \\

    \midrule
    \multicolumn{7}{l}{\textit{Audio LLM generation uncertainty}} \\

    \quad Phi-4 mean token probability
    & 80.80 & 82.26 & 0.19 
    & 79.75 & 80.69 & 0.20
    \\

    \quad Phi-4 mean token entropy
    & 81.93 & 83.44 & 0.17
    & 79.46 & 80.75 & 0.20
    \\

    \midrule
    \multicolumn{7}{l}{\textit{Transcript-conditioned WER estimation}} \\

    \quad Fe-WER~\citep{park2025fast}
    & 75.44 & 79.03 & 0.23 
    & 70.47 & 74.89 & 0.27
    \\

    \midrule
    \multicolumn{7}{l}{\textit{Audio-encoder reliability prediction (ours)}} \\

    \quad \textbf{Phi-4 audio encoder + reliability predictor}
    & \textbf{87.01} 
    & \textbf{88.02} 
    & \textbf{0.12}
    & \textbf{83.54} 
    & \textbf{84.35} 
    & \textbf{0.16}
    \\

    \bottomrule
    \end{tabular}
    }
\end{table}

\section{Additional Evaluation Results}
\label{app:additional_results}

Table~\ref{tab:phi4_main_results} shows that the trends observed for Qwen2-Audio-7B-Instruct in Table~\ref{tab:main_results} also hold for Phi-4-Multimodal-Instruct. Our reliability predictor achieves $87.01\%$ macro-F1, $88.02\%$ accuracy, and $0.12$ MAE on the in-domain test set. The strongest competing baseline, Phi-4 mean token entropy, reaches $81.93\%$ macro-F1, $83.44\%$ accuracy, and $0.17$ MAE. Our predictor therefore improves macro-F1 by $5.08$ points and accuracy by $4.58$ points while reducing MAE by $0.05$.

The predictor remains strong under cross-domain evaluation, achieving $83.54\%$ macro-F1, $84.35\%$ accuracy, and $0.16$ MAE, corresponding to drops of only $3.47$ and $3.67$ points in macro-F1 and accuracy relative to the in-domain setting. Among the baselines, Phi-4 mean token probability achieves the highest macro-F1 at $79.75\%$, while mean token entropy achieves the highest accuracy at $80.75\%$; both reach an MAE of $0.20$. Relative to the strongest baseline for each metric, our predictor improves macro-F1 by $3.79$ points and accuracy by $3.60$ points while reducing MAE by $0.04$.

Table~\ref{tab:moss_main_results} shows a similar pattern for MOSS-Audio-8B. Our reliability predictor achieves $82.97\%$ macro-F1, $84.44\%$ accuracy, and $0.16$ MAE on the in-domain test set. The strongest competing baseline, MOSS mean token entropy, reaches $74.83\%$ macro-F1, $77.95\%$ accuracy, and $0.24$ MAE. This corresponds to improvements of $8.14$ points in macro-F1 and $6.49$ points in accuracy, together with a $0.08$ reduction in MAE.

Under cross-domain evaluation, the predictor retains $81.60\%$ macro-F1, $82.86\%$ accuracy, and $0.17$ MAE, with only $1.37$- and $1.58$-point drops in macro-F1 and accuracy from the in-domain setting. MOSS mean token entropy is again the strongest baseline, achieving $76.77\%$ macro-F1, $78.78\%$ accuracy, and $0.22$ MAE. Our predictor improves these results by $4.83$ macro-F1 points and $4.08$ accuracy points while reducing MAE by $0.05$. Together with the Qwen2-Audio-7B-Instruct results in Table~\ref{tab:main_results}, these results show that audio-encoder reliability prediction consistently outperforms the evaluated baselines across all three Audio LLM families.

\begin{table}[t]
    \centering
    \caption{ 
    \textbf{Four-class transcription reliability prediction and cross-domain generalization for MOSS-Audio-8B.} We report macro-F1, accuracy, and mean absolute error (MAE) on the in-domain and cross-domain test sets, following the same baselines and evaluation protocol as the Qwen2-Audio-7B-Instruct results in Table~\ref{tab:main_results}. The proposed reliability predictor achieves the strongest overall performance in both settings. 
    } 
    \label{tab:moss_main_results}

    \resizebox{\linewidth}{!}{
    \begin{tabular}{lcccccc}
    \toprule
    &
    \multicolumn{3}{c}{\textbf{In-domain Test}}
    &
    \multicolumn{3}{c}{\textbf{Cross-domain Test}}
    \\
    \cmidrule(lr){2-4}
    \cmidrule(lr){5-7}

    \textbf{Method}
    & \textbf{Macro-F1} $\uparrow$
    & \textbf{Accuracy} $\uparrow$
    & \textbf{MAE} $\downarrow$
    & \textbf{Macro-F1} $\uparrow$
    & \textbf{Accuracy} $\uparrow$
    & \textbf{MAE} $\downarrow$
    \\
    \midrule

    \multicolumn{7}{l}{\textit{Acoustic corruption proxy}} \\

    \quad SNR
    & 55.43 & 67.77 & 0.36
    & 51.99 & 64.00 & 0.42
    \\

    \midrule
    \multicolumn{7}{l}{\textit{No-reference speech quality and intelligibility predictors}} \\

    \quad Audiobox-Aesthetics (PQ)~\citep{tjandra2025audioboxaesthetics}
    & 22.90 & 28.97 & 1.05 
    & 22.05 & 28.26 & 0.87
    \\

    \quad DNSMOS (OVRL)~\citep{reddy2021dnsmos}
    & 38.70 & 47.17 & 0.59
    & 42.17 & 52.08 & 0.57
    \\

    \quad NISQA (MOS)~\citep{mittag2021nisqa}
    & 56.12 & 58.55 & 0.44
    & 53.04 & 62.94 & 0.37
    \\

    \quad TorchAudio-SQUIM (STOI)~\citep{kumar2023squim}
    & 62.32 & 64.96 & 0.37   
    & 63.59 & 66.82 & 0.36
    \\

    \midrule
    \multicolumn{7}{l}{\textit{Audio LLM generation uncertainty}} \\

    \quad MOSS mean token probability
    & 73.47 & 76.29 & 0.27 
    & 74.93 & 76.16 & 0.26
    \\

    \quad MOSS mean token entropy
    & 74.83 & 77.95 & 0.24
    & 76.77 & 78.78 & 0.22
    \\

    \midrule
    \multicolumn{7}{l}{\textit{Transcript-conditioned WER estimation}} \\

    \quad Fe-WER~\citep{park2025fast}
    & 72.32 & 76.45 & 0.28
    & 69.19 & 74.14 & 0.29
    \\

    \midrule
    \multicolumn{7}{l}{\textit{Audio-encoder reliability prediction (ours)}} \\

    \quad \textbf{MOSS audio encoder + reliability predictor}
    & \textbf{82.97}
    & \textbf{84.44} 
    & \textbf{0.16}
    & \textbf{81.60} 
    & \textbf{82.86} 
    & \textbf{0.17}
    \\

    \bottomrule
    \end{tabular}
    }
\end{table}

\section{Confusion Matrix Analysis}
\label{app:confusion_matrix}

\begin{figure}[t]
    \centering
    \includegraphics[width=\linewidth]
    {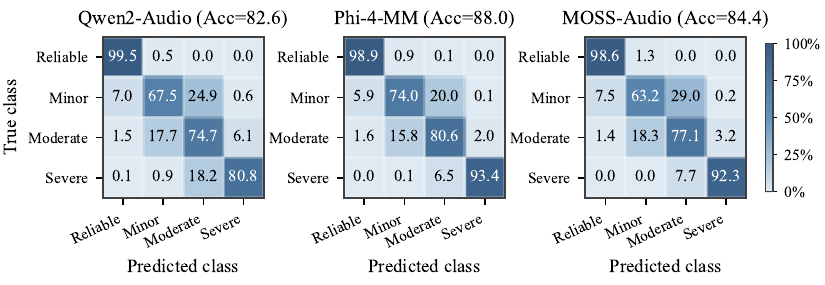}
    \caption{
    \textbf{Confusion matrices for transcription reliability prediction
    across target Audio LLMs.}
    Results are shown for Qwen2-Audio-7B-Instruct, Phi-4-Multimodal, and MOSS-Audio-8B. Rows represent the ground-truth reliability classes, columns represent the predicted reliability classes, and each cell reports the percentage of examples from the corresponding ground-truth class. The accuracy for each target Audio LLM is reported above its confusion matrix.
    }
    \label{fig:confusion_matrix}
\end{figure}

Figure~\ref{fig:confusion_matrix} shows the confusion matrices for the reliability predictors of the three target Audio LLMs. The predictors achieve accuracies of 82.6\%, 88.0\%, and 84.4\% for Qwen2-Audio-7B-Instruct, Phi-4-Multimodal, and MOSS-Audio-8B, respectively. Importantly, recall for the reliable class is 99.5\%, 98.9\%, and 98.6\%, respectively. High recall for this class prevents unnecessary clarification requests for audio that the target Audio LLM can transcribe without error. Across all three models, most incorrect predictions occur between neighboring reliability classes, particularly between minor and moderate degradation.

\section{Computational Overhead Analysis}
\label{app:efficiency}

\begin{table}[t]
    \centering
    \caption{
    \textbf{Computational overhead of the reliability predictor.}
    We report the parameter counts and FLOPs of the audio encoder, language model
    decoder, and proposed prediction head. FLOPs are measured with \texttt{torch.profiler} on an approximately 12-second input recording using a single forward pass. Methods relying on generated transcripts
    or generation uncertainty require executing the decoder, whereas our predictor
    operates directly on the encoder representations.
    }
    \label{tab:efficiency}
    \setlength{\tabcolsep}{2.3pt}
    \normalsize
    \begin{tabular}{@{}lcccccc@{}}
        \toprule
        & \multicolumn{3}{c}{Parameters}
        & \multicolumn{3}{c}{FLOPs} \\
        \cmidrule(lr){2-4}
        \cmidrule(lr){5-7}
        Model
        & Encoder
        & Decoder
        & Head
        & Encoder
        & Decoder
        & Head \\
        \midrule
        Qwen2-Audio-7B-Instruct
        & 636.97M
        & 7.76B
        & 0.33M
        & 1.89T
        & 4.21T
        & 2.12M \\

        Phi-4-Multimodal-Instruct
        & 441.24M
        & 3.84B
        & 0.26M
        & 114.20G
        & 1.32T
        & 1.70M \\

        MOSS-Audio-8B
        & 643.62M
        & 8.19B
        & 329.99K
        & 243.05G
        & 2.67T
        & 2.12M \\
        \bottomrule
    \end{tabular}
\end{table}

The proposed reliability predictor introduces minimal additional computation to the standard Audio LLM inference pipeline. Since the audio encoder is already executed to obtain representations for downstream generation, our method only adds a lightweight prediction head on top of the frozen encoder outputs. Importantly, the reliability decision can be made without executing the language model decoder.

This differs from reliability signals that depend on the generated output. Audio LLM generation uncertainty requires quantities produced during decoding, while transcript-conditioned WER estimation requires a generated transcript. Therefore, these approaches must first execute the Audio LLM decoder before reliability can be estimated. This cost is substantial relative to our prediction head, particularly when reliability is used to decide whether generation should proceed at all.

Table~\ref{tab:efficiency} summarizes the parameter counts and computational costs of the audio encoder, language model decoder, and proposed reliability predictor.
\footnote{The reported FLOPs are measured with \texttt{torch.profiler} on an approximately 12-second input recording using a single forward pass. The decoder FLOPs correspond to the prefill stage and exclude autoregressive generation, whose cost depends on output length. Since transcription outputs contain relatively few tokens, including generation would only modestly increase decoder cost and would not affect the conclusion that decoding is orders of magnitude more expensive than the proposed reliability predictor.}
Across all evaluated Audio LLMs, the prediction head is several orders of magnitude smaller than both the encoder and decoder. For example, for Qwen2-Audio-7B-Instruct, the predictor contains only 0.33M trainable parameters and requires 2.12M FLOPs, compared with 636.97M parameters and 1.89T FLOPs for the audio encoder and 7.76B parameters and 4.21T FLOPs for the decoder. Thus, our predictor can estimate transcription reliability immediately after audio encoding while avoiding the substantially larger decoding cost required by post-generation reliability methods.
\section{Ablation Study}
\label{app:ablation}

We conduct three ablation studies to better understand the design choices behind the proposed reliability predictor and cross-model label transfer. First, we compare different methods for aggregating the sequence of frozen audio-encoder representations into an utterance embedding using Qwen2-Audio-7B-Instruct. Second, we probe representations extracted from different depths of the audio encoders across three Audio LLMs to investigate where transcription reliability information emerges. Third, we ablate the strategy used to estimate shifts between the source and target critical-SNR boundaries.

\subsection{Temporal Aggregation.} 
\label{app:temporal_aggregation}
The audio encoder produces a sequence of temporally aligned representations, which must be summarized before reliability classification. We compare mean pooling, max pooling, and learned temporal pooling. In all cases, the Qwen2-Audio-7B-Instruct audio encoder remains frozen, and only the aggregation module and classification head are optimized. This comparison isolates the effect of temporal aggregation while keeping the underlying pretrained representations and training data fixed. As shown in Table~\ref{tab:pooling_ablation}, learned temporal pooling achieves the strongest overall performance, reaching 81.10\% macro-F1 and 82.60\% accuracy with an MAE of 0.18. Mean and max pooling perform worse, suggesting that uniformly aggregating the frame-level representations or retaining only their strongest activations discards information relevant to transcription reliability. These results indicate that transcription reliability benefits from learning how much weight to assign to different frames.

\begin{table}[t]
    \centering
    \caption{
    \textbf{Ablation of temporal aggregation methods for transcription
    reliability prediction.}
    All methods operate on frozen Qwen2-Audio-7B-Instruct audio-encoder
    representations and use the same transcription reliability dataset.
    Only the temporal aggregation module and classification head are trained.
    }
    \label{tab:pooling_ablation}
    \vspace{1mm}

    \setlength{\tabcolsep}{4pt}
    \begin{tabular}{lccc}
        \toprule
        Pooling Method
        & Macro-F1 $\uparrow$
        & Accuracy $\uparrow$
        & MAE $\downarrow$ \\
        \midrule
        Max Pooling
        & 77.60 & 79.50 & 0.22 \\
        Mean Pooling
        & 79.69 & 81.35 & 0.20 \\
        Learned Temporal Pooling
        & \textbf{81.10} & \textbf{82.60} & \textbf{0.18} \\
        \bottomrule
    \end{tabular}
\end{table}

\begin{table}[t]
    \centering
    \caption{
    \textbf{Ablation of critical-SNR alignment strategies for cross-model
    reliability label transfer.}
    We compare direct cross-model transfer with several strategies for
    aligning source-model critical-SNR boundaries to the target model using
    $5\%$ of the shared speech--corruption pairs. We report the accuracy drop
    relative to training with the target model's own label bank; lower is better.
    }
    \label{tab:shift_strategy_ablation}

    \resizebox{\linewidth}{!}{%
    \begin{tabular}{lccccc}
        \toprule
        Transfer setting
        & \multicolumn{1}{c}{Direct Transfer}
        & \multicolumn{4}{c}{Aligned Transfer} \\
        \cmidrule(lr){3-6}
        Alignment method
        &
        & Global Mean
        & Global Median
        & Per-Boundary Mean
        & Per-Boundary Median \\
        \midrule

        MOSS $\leftarrow$ Qwen2
        & 9.69
        & 1.90
        & 1.92
        & 2.30
        & \textbf{1.70} \\

        MOSS $\leftarrow$ Phi-4
        & 12.86
        & 2.71
        & 2.29
        & 2.21
        & \textbf{2.05} \\

        \midrule
        Mean $\downarrow$
        & 11.28
        & 2.31
        & 2.10
        & 2.26
        & \textbf{1.88} \\

        \bottomrule
    \end{tabular}%
    }
\end{table}

\subsection{Encoder Probe Depth.} 
\label{app:encoder_probe_depth}
We further investigate where transcription reliability information emerges within the audio encoder by probing representations extracted from different encoder depths. For each layer, we freeze the corresponding audio representation and train the same lightweight reliability predictor used in the main experiments.

\begin{figure}[t]
    \centering
        \vspace{0pt}
        \centering
        \includegraphics[width=\linewidth]
        {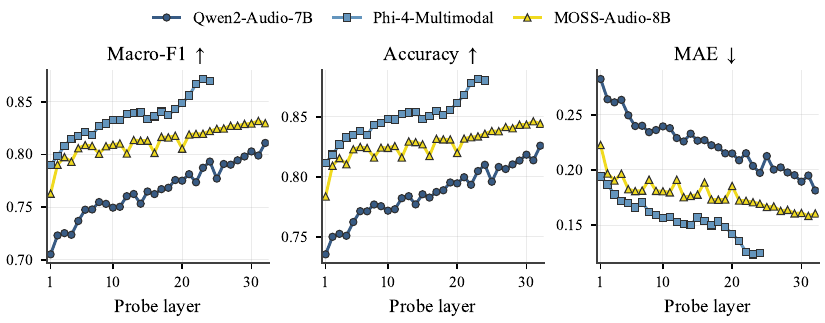}
        \caption{
    \textbf{Transcription reliability information becomes stronger in deeper audio-encoder layers.}
    We probe frozen representations extracted from increasing depths of the Qwen2-Audio-7B-Instruct, Phi-4-Multimodal-Instruct, and MOSS-Audio-8B audio encoders using the same lightweight reliability predictor. Deeper layers generally provide more informative representations, improving macro-F1 and accuracy while reducing MAE.
    }
        \label{fig:probe_layer_sweep}
\end{figure}

As shown in Figure~\ref{fig:probe_layer_sweep}, reliability prediction generally improves as we move toward deeper encoder layers. Across all evaluated Audio LLMs, deeper representations tend to achieve higher macro-F1 and accuracy while reducing MAE. This trend suggests that deeper audio-encoder representations capture increasingly informative features associated with model-specific transcription reliability.

\subsection{Critical-SNR Alignment Strategy.}
\label{app:shift_strategy_ablation}
We further ablate the strategy used to estimate shifts between the source and target critical-SNR boundaries. We compare global and per-boundary estimates using either the mean or median, along with direct transfer without alignment. As shown in Table~\ref{tab:shift_strategy_ablation}, all alignment strategies substantially improve over direct transfer. Among the evaluated variants, the per-boundary median achieves the strongest overall performance, yielding the lowest average accuracy drop across the two transfer settings. In principle, applying separate per-boundary shifts can invert the ordering of adjacent critical-SNR boundaries. We therefore exclude such invalid cases. In our experiments, these ordering violations were extremely rare.

\end{document}